\documentclass[sigconf,nonacm,screen=true]{acmart}

\title{Gesture-Based Robot Control Integrating Mm-wave Radar and Behavior Trees}

\author{Yuqing Song}
\authornotemark[1]
\email{yuqing.song@aalto.fi}

\affiliation{%
  \institution{Aalto University}
  \city{Espoo}
  \country{Finland}
}

\author{Cesare Tonola}
\authornote{Both authors contributed equally to this research.}
\email{cesare.tonola@cnr.it}
\affiliation{%
  \institution{National Research Council of Italy}
  \city{Milan}
  \country{Italy}
}

\author{Stefano Savazzi}
\email{stefano.savazzi@cnr.it}
\affiliation{%
  \institution{National Research Council of Italy}
  \city{Milan}
  \country{Italy}}

\author{Sanaz Kianoush}
\email{sanaz.kianoush@cnr.it}
\affiliation{%
  \institution{National Research Council of Italy}
  \city{Milan}
  \country{Italy}
}

\author{Nicola Pedrocchi}
\email{nicola.pedrocchi@stiima.cnr.it}
\affiliation{%
  \institution{National Research Council of Italy}
  \city{Milan}
  \country{Italy}
}

\author{Stephan Sigg}
\email{stephan.sigg@aalto.fi}
\affiliation{%
  \institution{Aalto University}
  \city{Espoo}
  \country{Finland}
}

\renewcommand{\shortauthors}{Yuqing Song et al.}

\begin{document}

\begin{abstract}
As robots become increasingly prevalent in both homes and industrial settings, the demand for intuitive and efficient human-machine interaction continues to rise. Gesture recognition offers an intuitive control method that does not require physical contact with devices and can be implemented using various sensing technologies. Wireless solutions are particularly flexible and minimally invasive. While camera-based vision systems are commonly used, they often raise privacy concerns and can struggle in complex or poorly lit environments. In contrast, radar sensing preserves privacy, is robust to occlusions and lighting, and provides rich spatial data such as distance, relative velocity, and angle.
We present a gesture-controlled robotic arm using mm-wave radar for reliable, contactless motion recognition. Nine gestures are recognized and mapped to real-time commands with precision. Case studies are conducted to demonstrate the system practicality, performance and reliability for gesture-based robotic manipulation. Unlike prior work that treats gesture recognition and robotic control separately, our system unifies both into a real-time pipeline for seamless, contactless human-robot interaction.

\end{abstract}



\begin{teaserfigure}
\centering
  \includegraphics[width=0.9\textwidth]{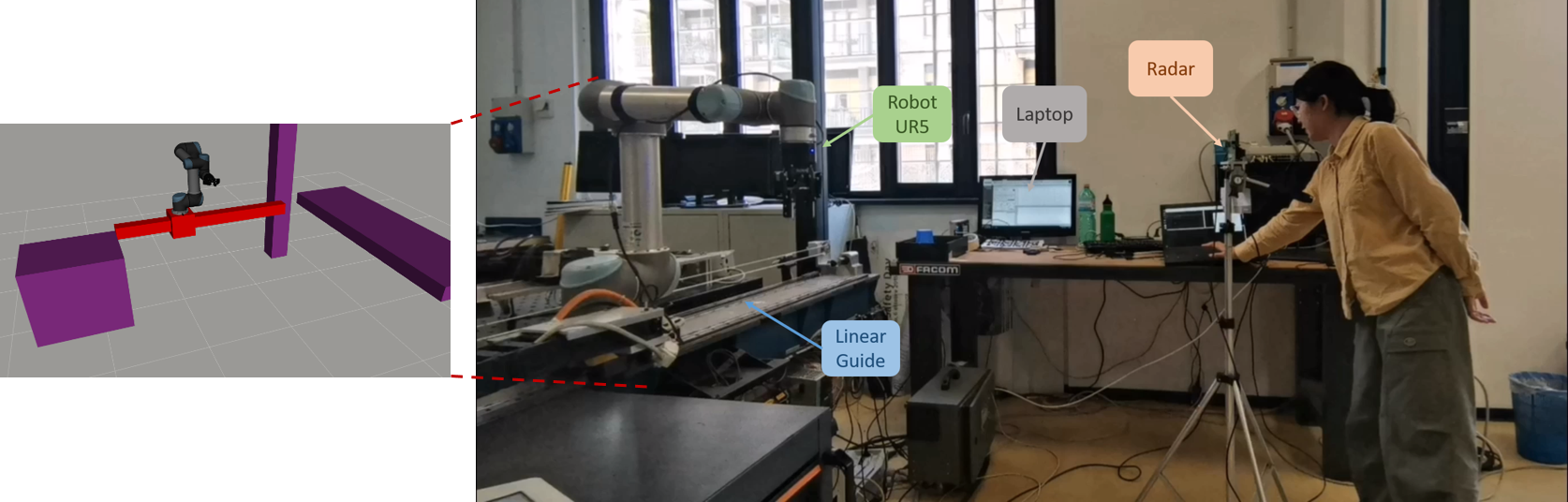}
  \caption{Experimental site of millimeter-wave radar-based gesture-controlled robotic arm. The inset on the left shows the digital twin of the robotic arm and linear guide used for on-site demonstration.}
  \Description{The image captures a live experiment demonstrating gesture recognition using mmWave radar to control a robotic arm. The participant stands in front of the radar and performs specific hand gestures, which are detected and interpreted in real time. Based on the recognized gestures, the robotic arm positioned behind the radar executes corresponding movements, enabling contactless human-robot interaction.}
  \label{fig:teaser}
\end{teaserfigure}


\maketitle

\section{Introduction}
Robotic arms have become increasingly important in recent years, especially in fields such as industrial automation \cite{sanjeev2025robotic}, healthcare \cite{kyrarini2021survey}, agriculture \cite{JIN2024108938}, and service robotics \cite{gonzalez2021service}. They support continuous operations, highly efficient, and capable of operating in hazardous environments that may be unsafe for humans. With their growing adoption in daily life, there is a pressing need for more intuitive and efficient ways to interact with them-especially in scenarios where full automation is neither feasible nor desirable, and human judgment remains essential.
For example, in assistive care \cite{sather2021assistive}, elderly or disabled individuals intuitive and accessible means to control a robotic arm for tasks like feeding, picking up objects, delivering, and lifting. In medical settings, healthcare personnel can rely on teleoperated robots \cite{silvera2024robotics} to adjust cameras or lighting equipment and deliver medicine or tools, all without physical contact, ensuring sterile operation. In hazardous environments such as areas with toxic chemicals, high temperatures, or radiation \cite{trevelyan2016robotics}, it becomes essential for human operators to remain in safe zones while controlling robotic arms remotely to perform critical tasks.

Common ways to control robotic arms include physical devices \cite{rahman2019joystick} such as buttons, joysticks, or remotes. These methods are mature and reliable, and they provide precise control in many applications. However, they may not be intuitive for general users and often require a certain level of training. Touchscreen interfaces \cite{rabhi2018intelligent} offer a more visual and flexible way to interact with robotic systems, but they still require manual operation and sustained attention, which may not be suitable for users without computer skills. Voice control \cite{kanash2021design, 9687192} provides a contactless alternative, yet it can be affected by noise, accents, and privacy concerns.
Gesture recognition \cite{bularka2018robotic} offers a natural and contactless solution for a wide range of users, providing a seamless human-machine interaction experience. This makes it especially useful in situations where hands may be occupied, hygiene is important, or quick reactions are needed. Moreover, compared to other modalities like images, radar-based gesture recognition brings unique advantages such as enhanced privacy protection, robustness to environmental variations, and reliable operation under poor lighting or occlusion \cite{ahmed2021hand}.

\subsection{Contributions}

In summary, this work presents a gesture-controlled robotic arm integrating millimeter-wave (mm-wave) radar and behavior trees. The main contributions of this work are:

\begin{itemize}
\item \textbf{End-to-end integration of gesture recognition and robotic control:} Unlike most radar-based gesture recognition research \cite{yu2024mmwave, ahmed2021hand, liu2023multimodal, zhang2020temporal, bhavanasi2022patient, wang2020hand, salami2022tesla} that focuses only on classification, we develop a complete system where nine distinct hand gestures recognized by a robust, contactless mm-wave radar pipeline are translated into real-time commands. By employing behavior trees to define robot tasks triggered by gestures, our system allows intuitive human-robot interaction and easy customization of robot behaviors without requiring specialized programming expertise.

\item \textbf{Practical demonstration through real-world case studies:} We validate the system's effectiveness and reliability via several tasks, including picking up, moving, and placing a cup, pouring water, and executing emergency stops, fully controlled by hand gestures.
\end{itemize}

\section{Methods}

This part involves (1) the proposed architecture of gesture recognition in Section 3.1, (2)  hardware architecture in Section 3.2, and (3) overall software architecture in Section 3.3.

\subsection{Gesture Recognition – Proposed Architecture} \label{sec: gesture_recognition}

\begin{figure}[t]
  \centering
  \includegraphics[width=0.9\columnwidth]{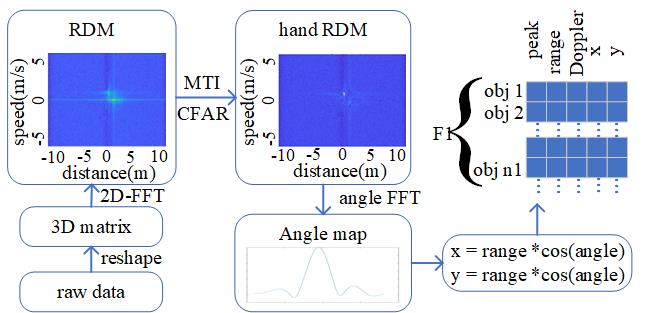}
  \caption{Radar data processing methods.}
  \label{fig:pro}
\end{figure}

As depicted in Fig \ref{fig:pro}, the data processing pipeline converts raw radar input into the point cloud representation. Starting from a 3D matrix of size $N_c \times N_{\text{chirp}} \times N_{\text{chan}}$ (number of channels × \text{number of chirps} × \text{number of antenna elements}), a 2D-FFT is applied per channel and summed into an RDM. MTI (Moving Target Indication) and CFAR are applied to filter out static clutter and isolate moving gesture targets. An FFT across the $N_{\text{chan}}$ dimension yields the angle map (angle-amplitude map), from which Cartesian coordinates are calculated. Each frame outputs a list of detected objects with peak, range (distance), Doppler, x, and y features for further analysis. The complete processing chain is implemented on-chip and runs on the radar SoC (System on Chip).


\begin{figure}[t]
  \centering
  \includegraphics[width=0.8\columnwidth]{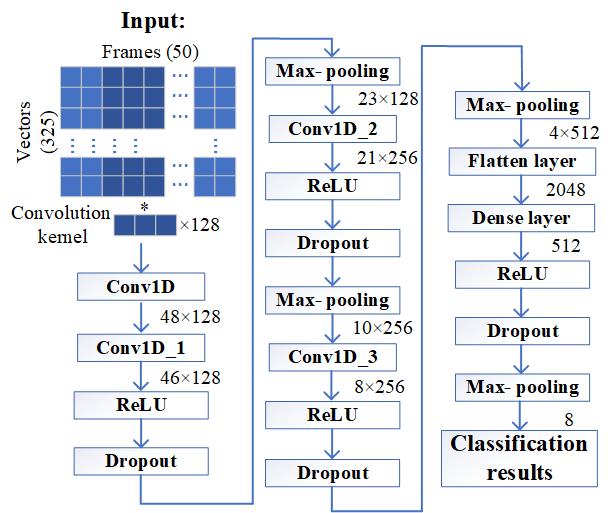}
  \caption{1D-CNN for gesture recognition.}
  \label{fig:cnn}
\end{figure}

To simplify the algorithm and accelerate the gesture recognition process, a one-dimensional Convolutional Neural Network (CNN) is applied, which is illustrated in Fig \ref{fig:cnn}. The maximum number of frames is 50, and the maximum number of detected objects per frame is 65, resulting in each gesture sample being represented as a 50 × 325 matrix. The input tensor (50, 325) first passes through two consecutive Conv1D layers with 128 filters (kernel size 3), reducing the shape to (46, 128), followed by ReLU, dropout, and max pooling, downsampling the time dimension to 23. A third Conv1D layer with 256 filters (size 3) and a fourth with 512 filters further compress the feature map to (4, 512) through similar post-processing. The output is then flattened to a 2048-dimensional vector, mapped to 512 units via a dense layer with ReLU and dropout, and finally to an 8-dimensional vector representing gesture class logits. The entire process is implemented in real time using buffer management. 

\subsection{Hardware Architecture}

As shown in Fig~\ref{fig:teaser}, the hardware setup includes a UR5 robot with a Robotiq 2f-85 gripper mounted on a linear guide for extended motion. The 6-DoF (Degrees of Freedom) UR5, combined with the linear guide, increases the robot's workspace. A laptop controls the entire system, and an AWR1642 radar with 2 TX and 4 RX antennas is used for gesture recognition.


\subsection{Overall Software Architecture} \label{sec: sw architecture}

\begin{figure}[t]
  \centering
  \includegraphics[width=\columnwidth]{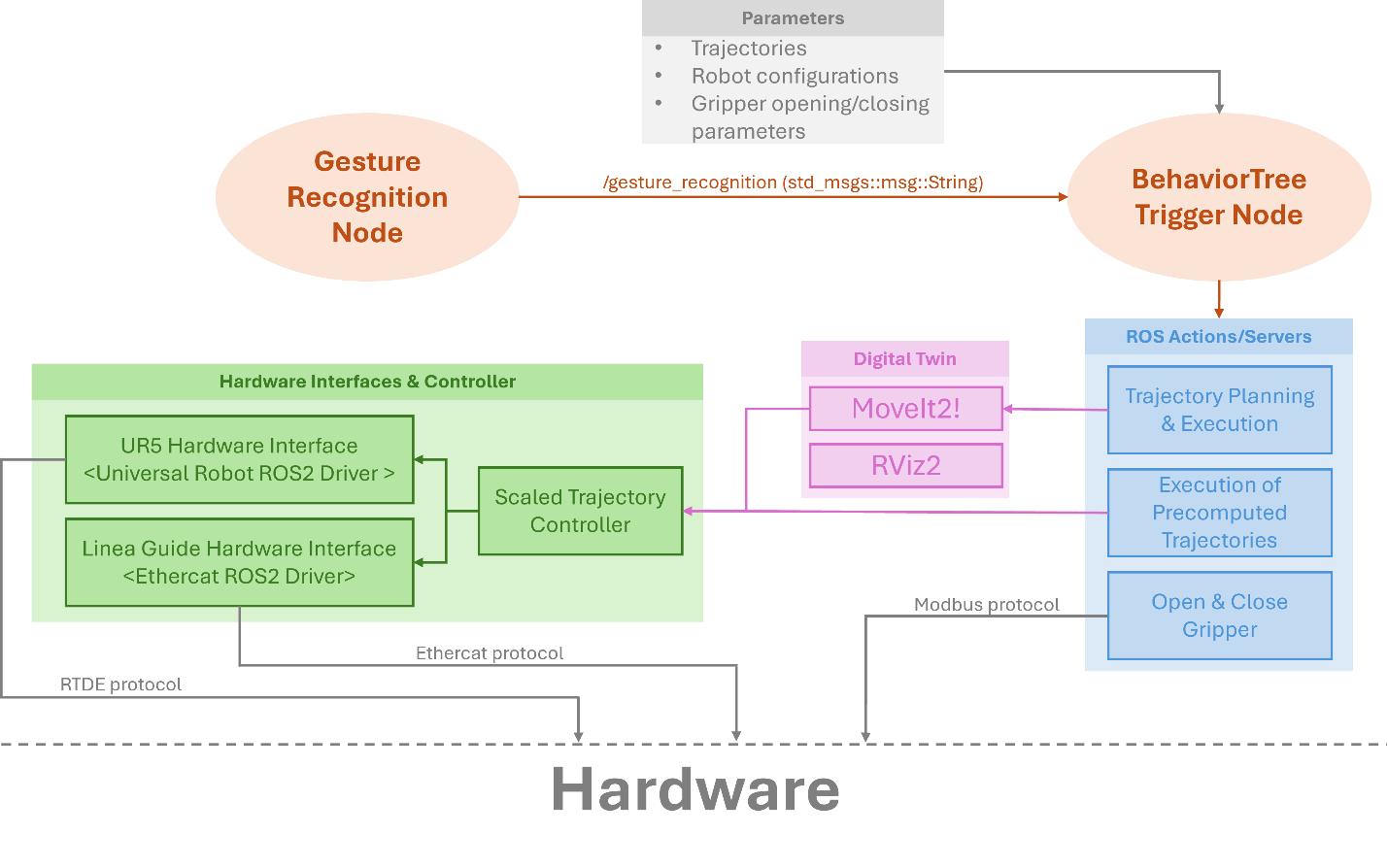}
  \caption{Overall software architecture.}
  \label{fig:software}
\end{figure}

The robotic cell in Fig \ref{fig:software} simulates an industrial scenario where a human operator controls tasks using radar-based gestures, typically a pick-and-place operation. Each gesture triggers a behavior tree, representing either a full task or a single action, depending on the control level. The operator completes the task by performing the correct gesture sequence.


The entire control stack is based on the third-party software frameworks and libraries listed in Table \ref{tab:ros2_libraries}.

\begin{table*}[htp]
\centering
\renewcommand{\arraystretch}{1.2}  
\begin{tabular}{p{5cm} p{11cm}}
\toprule
\textbf{Library} & \textbf{Description} \\
\midrule
\textbf{ROS 2}~\cite{macenski2022robot, doi:10.48550/arXiv.2305.09933} & Middleware for modular development of robotic applications. Enables communication between software components (nodes) through \textit{topics}, \textit{services}, and \textit{actions}. \\
\textbf{ros2\_control}~\cite{ROS2_control} & ROS 2 framework for real-time robot control. Uses hardware interfaces to read joint states and send commands, connecting physical hardware with control logic. \\
\textbf{BehaviorTree.CPP}~\cite{bt} & C++ library for building and executing behavior trees. \\
\textbf{MoveIt 2}~\cite{moveit} & ROS 2 platform for motion planning, manipulation, and control. \\
\bottomrule
\end{tabular}
\caption{Libraries and frameworks used in proposed architecture.}
\label{tab:ros2_libraries}
\end{table*}

The architecture includes a dedicated ROS 2 node, \textit{Gesture Recognition Node}, which processes radar data to detect gestures using the algorithm in Section \ref{sec: gesture_recognition}. It publishes a string identifier to the topic \texttt{/gesture\_recognition}. Another node, \textit{BehaviorTree Trigger Node}, subscribes to this topic and triggers the corresponding behavior tree using the BehaviorTree.CPP library. Each tree can represent a single skill or a sequence for complex tasks. These skills typically involve calling ROS services or actions that plan a trajectory to reach a desired robot configuration or pose, execute precomputed trajectories and open/close the robot gripper.
%
%
%
The skill set can be extended to include features like impedance control, force/torque control, or visual servoing for better adaptability. Each service/action is parameterized based on the task (\textit{e.g.}, target pose or gripper state). Trajectories are planned online via MoveIt 2 using a digital twin shown in Fig \ref{fig:teaser} for collision-free motion. A controller executes the trajectory with fine interpolation and sends real-time commands to the hardware interface. It also supports dynamic speed scaling from 0\% (full stop) to 100\% (nominal trajectory speed) based on human proximity, ensuring compliance with ISO/TS 15066 \cite{ISOTS15066}.


\section{Experimental Activities}

In this part, the setup is presented, 5 case studies are explored under this architecture, and the results are given.

\subsection{Experimental Setup}

The initial dataset, adopted from previous experiments, consisted of nine gesture classes, with approximately 2,000 samples per class. These data were collected under highly controlled laboratory conditions, where only the hand was positioned above the radar sensor and background interference was minimized. Although the training performance appeared very promising, during field testing, significant interference from nearby humans and electronic devices was observed when the radar was deployed in a real-world environment. 

Therefore, additional data were collected under realistic conditions and incorporated into the dataset. 5 participants were involved. The composition of the dataset is shown in Table \ref{tab:data}. Firstly, 30\% of the data collected in the noisy environment is used as the test set. The other data was divided into training and validation sets in a ratio of 7:3. 

\begin{table}
  \caption{Composition of the dataset}
  \label{tab:data}
  \begin{tabular}{cc}
    \toprule
    Settings&Samples\\
    \midrule
    only hand in front of the radar & $\approx2000$/class\\
    hand and human in front of the radar & $\approx200$/class\\
    robotic arm at the back of the human & $\approx200$/class\\
  \bottomrule
\end{tabular}
\end{table}

\subsection{Case Studies}
Based on the hardware and processing infrastructure previously described, five different tests have been carried out:
\begin{itemize}
\item \textbf{Test 1:} The human performs specific gestures to activate each individual robot action required to complete a full pick-and-place task with a glass.
\item \textbf{Test 2:} Same as test 1, but the robotic arm is behind the human, introducing interference.
\item \textbf{Test 3:} The human uses gestures to trigger a sequence of robot actions that pour water from a bottle into a glass.
\item \textbf{Test 4:} The human uses the "S" gesture to emergency stop the robotic arm. 
\item \textbf{Test 5:} The human continuously controls the positioning of the linear guide through hand movements (velocity control).
\end{itemize}

More specifically, in Tests 1 and 2, the gesture sequence was as follows:
\begin{itemize}
\item Performing the "swipe right" gesture moves the robotic arm to the right.
\item "Swipe CCW (counterclockwise)" opens the gripper.
\item "Down" moves the arm downward.
\item "Swipe CW (clockwise)" closes the gripper.
\item The arm then moves to the right position again.
\item "Swipe left" moves the arm to the left.
\item The "S" gesture triggers the object placement operation.
\item The arm then moves to the left position again.
\item Finally, the "up" gesture commands the robotic arm to return to its home position.
\end{itemize}

In Test 3, a more complex task involving multiple objects was performed:
\begin{itemize}
\item The "swipe right" gesture moves the robotic arm to the right.
\item The "down" gesture initiates the grasping of a glass.
\item "Swipe left" moves the arm to the left.
\item The "Z" gesture places the glass on the table.
\item The arm then moves right again.
\item The "X" gesture commands the arm to pick up a bottle.
\item After that, it moves left once more.
\item The "Swipe CW" gesture places the bottle next to the glass.
\item The "Swipe CCW" gesture triggers the pouring action to pour water into the glass.
\item Finally, the "up" gesture returns the robotic arm to the home position.
\end{itemize}

In Test 4, the emergency stop is achieved by rapidly scaling the robot's velocity down to zero using the trajectory controller described in Section \ref{sec: sw architecture}.
The velocity control logic in Test 5 operates as follows: each recognized hand gesture corresponds to a specific robot velocity command (for example, moving the linear axis left or right at a predefined speed). 
More specifically, the commanded velocity $v(t)$ is determined by the rule:

\begin{equation}
\begin{cases}
v(t) = v_{\mathrm{nom}} \cdot \lambda(t) \\
\lambda(t) = 0.5 \cdot \left(1 + \cos\left(\pi \frac{\min(t, T)}{T}\right)\right)
\end{cases}
\label{eq: velocity control}
\end{equation}
where $t$ is the elapsed time since the gesture was recognized, $v_{\mathrm{nom}}$ is the nominal velocity associated with the gesture, $\lambda(t)$ is a scaling factor that smoothly decreases the velocity to zero over time, and  $T$ is the duration after which the velocity reaches zero.
Each time a gesture is detected, the system selects the corresponding  $v_{\mathrm{nom}}$ and resets $t$ to zero. As time progresses without new gesture input, the velocity gradually decreases to zero according to Eq.~\ref{eq: velocity control}, ensuring safety. However, if the gesture is continuously maintained, the velocity remains close to $v_{\mathrm{nom}}$.

\subsection{Experimental Results}

\begin{figure}[t]
  \centering
  \includegraphics[width=1\columnwidth]{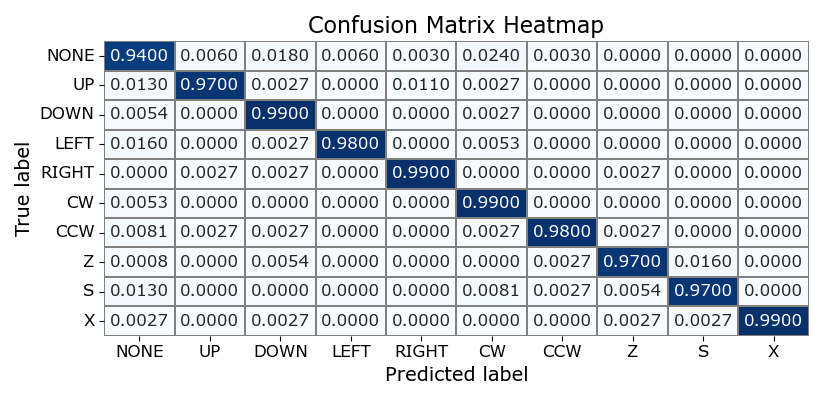}
  \caption{Confusion matrix of gesture recognition.}
  \label{fig:cm}
\end{figure}

After training the model with the updated data, the test results of the data collected in the field improved significantly, indicating that the gesture recognition system was robust and ready for practical deployment. The validation confusion matrix is shown in Fig \ref{fig:cm}. And the evaluation metrics of test set are as follows: Accuracy: 0.9368, Recall: 0.8426, F1-score: 0.8544. All case studies were successfully completed, and the interference in Test 2 did not influence at all. A demonstration video was recorded for each scenario.

\section{Conclusion and Future Work}

Our system uses mmWave radar to recognize hand gestures for touchless robotic arm control. Gestures trigger robot actions via behavior trees, offering robustness and privacy even in noisy environments.
While the current system only triggers robot actions through a codified gesture set, future developments will focus on enabling seamless, real-time control, allowing users to intuitively guide the robot's movements and operating range with enhanced freedom and precision. This could be achieved by expanding the number of recognizable gestures and enhancing the control logic. Given the nature of gesture control, we will also apply it to smart home scenarios such as music playback and lighting control.

\begin{acks}
This work was supported by EU Project HOLDEN, grant agreement 101099491.
\end{acks}

\bibliographystyle{unsrt}
\bibliography{sample-base}

\end{document}